\documentclass[11pt]{article}

\usepackage[preprint]{acl}

\usepackage{times}
\usepackage{latexsym}
\usepackage{booktabs} 
\usepackage{multirow} 
\usepackage{graphicx} 
\usepackage{rotating}   
\usepackage[most]{tcolorbox}
\usepackage[table]{xcolor}
\usepackage{graphicx}
\usepackage{subcaption}
\usepackage{enumitem}

\usepackage[T1]{fontenc}

\usepackage[utf8]{inputenc}

\usepackage{microtype}

\usepackage{inconsolata}

\usepackage{graphicx}
\usepackage{amsmath}
\usepackage{amssymb}
\usepackage{tabularx}
\usepackage{xcolor}
\definecolor{f1Green}{RGB}{138, 169, 113}

\title{MMedExpert-R1: Strengthening Multimodal Medical Reasoning via Domain-Specific Adaptation and Clinical Guideline Reinforcement}

\author{
  Meidan Ding$^{1,2,3,\dag}$, 
  Jipeng Zhang$^{4,\dag}$, 
  Wenxuan Wang$^{5}$,  
  \textbf{Haiqin Zhong}$^{6}$, \\
  \textbf{Xiaoling Luo}$^{1}$
  \textbf{Wenting Chen}$^{7}$, 
  \textbf{Linlin Shen}$^{1,2,3}$ \\
  $^{1}$College of Computer Science and Software Engineering, Shenzhen University \\
  $^{2}$School of Artificial Intelligence, Shenzhen University
  $^{7}$Stanford University \\
  $^{3}$Guangdong Provincial Key Laboratory of Intelligent Information Processing \\
  $^{4}$The Hong Kong University of Science and Technology 
  $^{5}$Renmin University of China\\
  $^{6}$School of Biomedical Engineering, Shenzhen University
}

\begin{document}
\maketitle
\begin{abstract}

Medical Vision-Language Models (MedVLMs) excel at perception tasks but struggle with complex clinical reasoning required in real-world scenarios. While reinforcement learning (RL) has been explored to enhance reasoning capabilities, existing approaches face critical mismatches: the scarcity of deep reasoning data, cold-start limits multi-specialty alignment, and standard RL algorithms fail to model clinical reasoning diversity. We propose MMedExpert-R1, a novel reasoning MedVLM that addresses these challenges through domain-specific adaptation and clinical guideline reinforcement. We construct MMedExpert, a high-quality dataset of 10K samples across four specialties with step-by-step reasoning traces. Our Domain-Specific Adaptation (DSA) creates specialty-specific LoRA modules to provide diverse initialization, while Guideline-Based Advantages (GBA) explicitly models different clinical reasoning perspectives to align with real-world diagnostic strategies. Conflict-Aware Capability Integration then merges these specialized experts into a unified agent, ensuring robust multi-specialty alignment. Comprehensive experiments demonstrate state-of-the-art performance, with our 7B model achieving 27.50 on MedXpert-MM and 83.03 on OmniMedVQA, establishing a robust foundation for reliable multimodal medical reasoning systems.

\end{abstract}

\section{Introduction}


\begin{figure}[t]
  \includegraphics[width=\columnwidth]{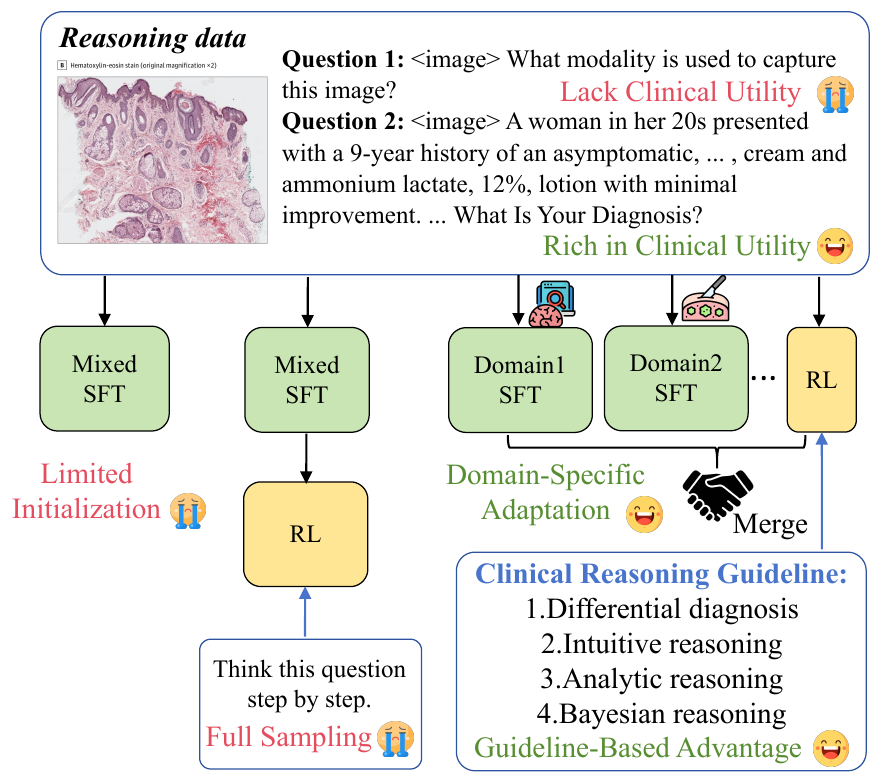}
  \caption{The core elements in MMedExpert-R1. The framework combines \textcolor{f1Green}{\textbf{Domain-Specific Adaptation (DSA)}} through specialty-aware LoRA modules, \textcolor{f1Green}{\textbf{Clinical Reasoning Data}} from mixed medical cases across multiple domains, and \textcolor{f1Green}{\textbf{Guideline-Based Advantage}} estimation guided by clinical reasoning guidelines. }
  
  \label{fig:introduction}
\end{figure}

Medical Vision-Language Models (MedVLMs) have proven effective in medical analysis~\cite{acosta2022multimodal} and assistive diagnosis~\cite{lipkova2022artificial}, achieving promising results in tasks such as medical report generation~\cite{che2025llm} and disease classification~\cite{jin2024health}. 
However, deploying these models in complex clinical scenarios often yields suboptimal performance. This stems from a critical capability gap: while existing MedVLMs are predominantly perception-oriented, they lack the sophisticated reasoning capabilities required to handle real-world clinical contexts. 
Specifically, clinical decision-making necessitates reasoning over diverse and interdependent information sources, synthesizing longitudinal patient data, and incorporating contextual medical knowledge, which remains a challenge for current perception-based approaches~\cite{su2025gmai}.

To overcome this limitation, reinforcement learning (RL) methods have been explored to bridge the gap between perception and reasoning, following their proven success in general scenarios like math~\cite{liu2025visual} and code generation~\cite{yang2025code}. 
A prominent stream involves leveraging RL algorithms, such as GRPO~\cite{guo2025deepseek}, during post-training to foster Chain of Thought (CoT) reasoning and self-reflection (e.g., Med-R1~\cite{lai2025med} and MedVLM-R1~\cite{pan2025medvlm}). 
These methods optimize the model by maximizing rule-based rewards, thereby explicitly encouraging robust clinical reasoning.

Though satisfactory performance, there are still \textbf{some mismatches} in reasoning MedVLMs with the elements of RL~\cite{lai2025med,pan2025medvlm,su2025gmai}, as shown in Figure~\ref{fig:introduction}. 

\noindent\textbf{Firstly, the difficulty of training data does not match clinical needs}. Current multimodal medical datasets~\cite{chen2024huatuogpt,zhang2023pmc} predominantly involve visual perception and shallow understanding tasks, whereas clinical questions demand step-wise differential diagnosis and evidence synthesis. In the elements of RL, the lack of deep reasoning supervision not only undermines data effectiveness but also prevents the reward signals from faithfully reflecting the true value of clinical reasoning. These observations motivate us to collect high-quality clinical data, which includes images, clinical records, and patient histories.


\noindent\textbf{Secondly, the RL with cold start limits multi-specialty alignment.} 
The standard RL pipeline is typically initialized from a single Supervised Fine-Tuning (SFT) checkpoint, effectively functioning as behavior cloning~\cite{su2025gmai,huang2025medvlthinker}. This paradigm yields a low-entropy policy that rigidly mimics the narrow distribution of the training data. In medicine, where clinical scenarios span diverse sub-specialties, this limitation is particularly severe: training data should cover varied specialties, yet the constrained initialization creates a "cold start" problem that restricts generation of diverse reasoning candidates. Thus, it is necessary to diversify the model distribution to expand the policy's exploration horizon and ensure robust alignment across distinct clinical paradigms.



\noindent \textbf{Thirdly, naive adoption of RL algorithms (e.g., GRPO) lacks medical domain modeling}. Although GRPO achieves stable optimization in general reasoning tasks, it shows clear limitations in medical scenarios. Its full-sampling assumption treats all reasoning trajectories as independent and equivalent, ignoring the diverse and complementary perspectives involved in real-world clinical decision-making. This clinical reasoning diversity—a core feature of medical cognition—should be explicitly modeled in RL sampling. 
In the elements of RL, the advantage function fails to differentiate reasoning quality, leading the model to overfit a single thinking pattern. Thus, it is essential to incorporate diverse reasoning paths into the RL process to align the model with real-world expert logic.


\noindent \textbf{Overall}, the mismatch across RL elements results in unstable optimization and poor generalization to real-world clinical reasoning. Addressing these systemic limitations is critical for developing reliable and clinically grounded multimodal reasoning models.

To overcome the aforementioned mismatches, we propose \textbf{MMedExpert-R1}, a novel reasoning MedVLM designed to cultivate reliable clinical reasoning by systematically enhancing the core elements of Reinforcement Learning, as illustrated in Figure~\ref{fig:introduction}. 
To bridge the data complexity gap, we \textbf{construct a high-quality medical reasoning dataset \textit{MMedExpert}} with 3.9K samples from four key specialties. Unlike standard VQA pairs, each entry couples a clinical scenario with a rigorous, step-by-step reasoning trace, providing the dense supervision required to model expert cognition. 
Based on this dataset, to provide multi-specialty model initialization, we propose a \textbf{Domain-Specific Adaptation (DSA) paradigm} to circumvent the exploration bottleneck inherent in standard checkpoints. DSA builds upon domain-aware LoRA modules, each fine-tuned on a single-specialty data, yielding expert models for each specialty. 
These modules recalibrate the initial policy distribution to cover a wider semantic manifold, preventing the agent from collapsing into suboptimal modes and providing a navigable search space for subsequent RL optimization.
Meanwhile, to align optimization with the nuance of real-world clinical decision-making, we introduce a \textbf{Guideline-Based Advantages (GBA)} module. This mechanism explicitly models diverse reasoning perspectives (e.g., differential vs. intuitive) with the advantages based on different clinical guidelines, enabling the policy to differentiate between high-quality diagnostic strategies and formulaic text generation rigorously.
For the multi-specialty alignment issue of RL with cold start, we propose a \textbf{Conflict-Aware Capability Integration} to finally merge the experts models from DSA and GBA into a unified reasoning agent to synthesize the dispersed capabilities.
MMedExpert-R1 enhances core RL components—training data quality, checkpoint design, and the advantage function—enabling more effective RL in medical contexts and producing clearer, clinically grounded reasoning. Extensive experiments demonstrate the superiority of MMedExpert-R1 to current LLMs across four datasets.
Overall, our contributions are as follows:
\begin{itemize}
\item We introduce MMedExpert, a high-quality, clinical reasoning dataset enriched with structured, step-by-step reasoning traces to support deeper and more interpretable clinical cognition.
\item We propose MMedExpert-R1, a novel reasoning MedVLM that cultivates specialized and reliable clinical reasoning by refining key elements of reinforcement learning.

\item We propose Domain-Specific Adaptation (DSA) and Guideline-Based Advantages (GBA) to improve RL-based medical reasoning, addressing limited generalization through domain-grounded initialization and enhancing reasoning diversity via guideline-based advantages.

\item Extensive experiments demonstrate the superiority of our MMedExpert-R1 to four existing medical benchmarks.


\end{itemize}

\section{Related Work}
\subsection{Medical Vision-Language Model }
The rapid development of Vision-Language Model (VLM) has led to remarkable advances across diverse domains, generating substantial interest in their application to the medical field~\cite{tian2023role,alsaad2024multimodal}. 
Early studies sought to integrate LLMs with specialized medical vision encoders, typically through a linear transformation layer, in order to achieve vision–language alignment for medical image understanding and analysis~\cite{li2023llava,moor2023med,liu2023qilin,zhang2024generalist}. Building upon this foundation, subsequent research has largely retained similar architectural paradigms while introducing a variety of strategies to improve performance. These strategies include constructing more comprehensive training datasets~\cite{ikezogwo2023quilt,chen2024huatuogpt,li2024gmai,hamamci2024developing}, designing sophisticated training recipes~\cite{nath2025vila,wang2024interactive}, employing efficient fine-tuning techniques~\cite{lin2025healthgpt}, incorporating mixture-of-experts mechanisms~\cite{he2024gsco}, and leveraging reinforcement learning~\cite{lai2025med,pan2025medvlm}.  
Building upon these developments, our work presents MMedExpert-R1, which incorporates domain-grounded initialization and clinical guideline reinforcement, enhancing both diagnostic reasoning and medical reliability.



\subsection{RL in MedVLMs} 
Recent studies have applied reinforcement learning (RL) to enhance medical reasoning, leading to significant gains in accuracy, interpretability, and generalization. Models such as Med-R1~\cite{lai2025med} and MedVLM-R1~\cite{pan2025medvlm} leverage GRPO to refine reasoning trajectories rather than relying solely on supervised fine-tuning, enabling more reliable decision-making across complex medical imaging tasks.  Similarly, GMAI-VL-R1~\cite{su2025gmai} integrates RL-based tuning with multi-modal data fusion, strengthening chain-of-thought reasoning and clinical robustness. 
Despite this progress, current MedVLMs still exhibit inconsistencies between their medical reasoning processes and the core elements of reinforcement learning. These structural mismatches hinder the full potential of RL in medical domains. Therefore, we aim to cultivate specialized, robust, and trustworthy clinical reasoning by systematically refining the key components of the RL framework for MedVLMs.

\subsection{Model Merging}

Model merging integrates complementary capabilities from multiple source models, yielding a single unified network directly within the parameter space. As a parameter-efficient optimization problem, merging algorithms calibrate the weight coefficients or resolve parameter interference to encode diverse functional skills, considering both task-specific knowledge and general representations.
Recently, model merging has been extended to cross-modal integration and heterogeneous architecture fusion to model complex reasoning capabilities.~\cite{yang2024model}.  ~\citet{chen2025bring} proposes merging models across modalities, effectively enabling the incorporation of the superior reasoning capabilities of LLMs into VLMs. ~\citet{zhang2024unconstrained} further proposes an unconstrained framework that accommodates both homogeneous and heterogeneous architectures.
Distinct from the aforementioned heterogeneous model fusion and only-SFT checkpoints fusion, we define the integration of models derived from knowledge-based SFT models and logic-based model RL as a unified merging task. This approach effectively harmonizes domain-specific medical knowledge with rigorous reasoning alignment, bridging the gap between raw professional expertise and preference-driven clinical decision-making.



\begin{figure*}[t]
  \includegraphics[width=\textwidth]{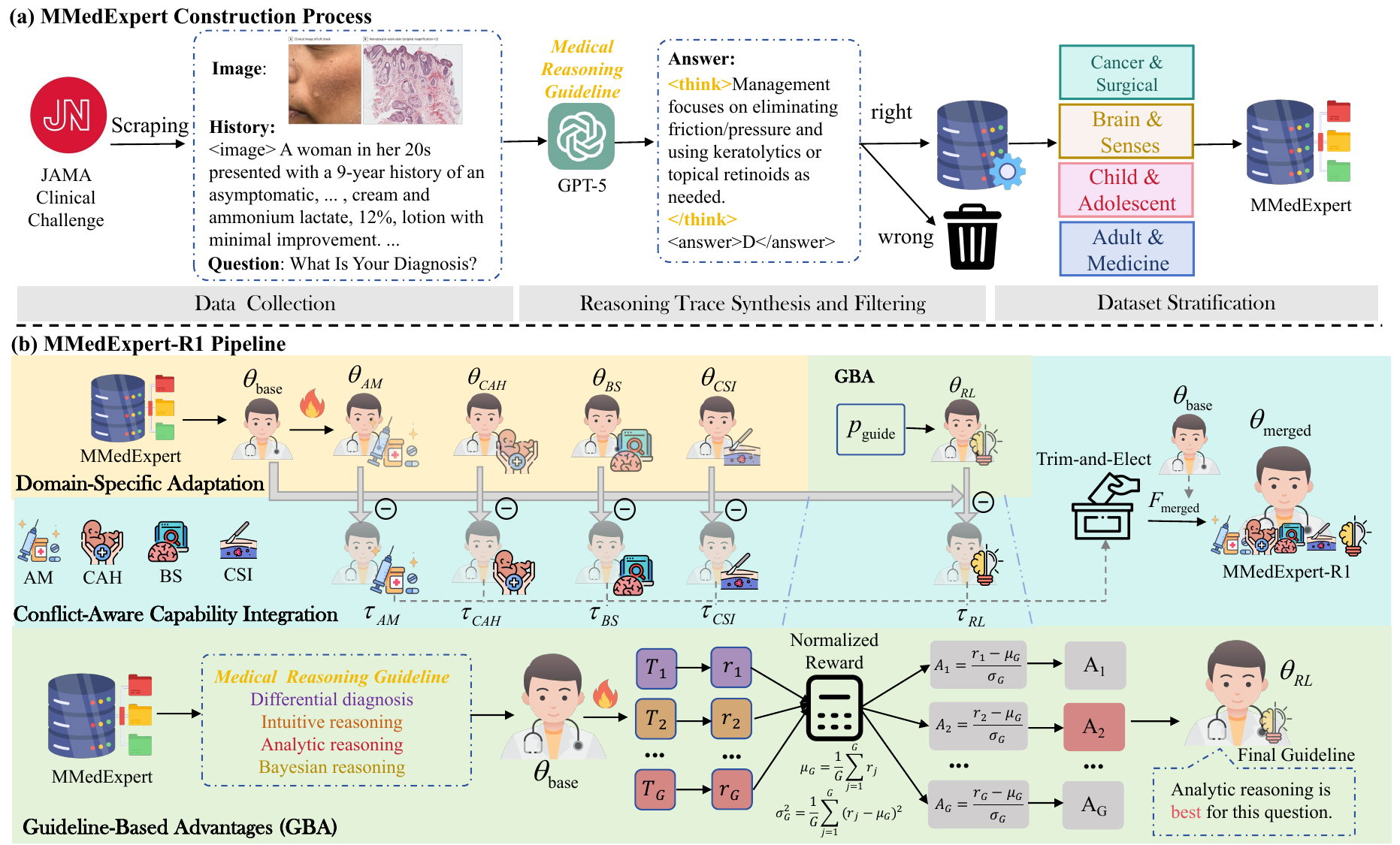}
  \caption{(a) Data construction of \textbf{MMedExpert}; (b) \textbf{MMedExpert-R1} consists of a Domain-Specific Adaptation to provide diverse initialization, Guideline-Based Advantages to align with real-world diagnostic strategies, and a Conflict-Aware Capability Integration to ensure multi-specialty alignment.
}
  \label{fig:method}
\end{figure*}

\section{MMedExpert}

Existing medical Visual Question Answering (VQA) datasets frequently suffer from a complexity gap, lacking the depth required to mirror decision-making processes in real-world clinical settings. To bridge this gap, we introduce \textit{MMedExpert}, a dataset explicitly synthesized to foster high-quality multimodal medical reasoning. Our construction pipeline proceeds in three rigorous stages.

\subsection{Data Collection}
To establish a robust foundation, we curated a source corpus of 10K raw samples (comprising 8K text-only and 2K multimodal instances) from authoritative clinical databases, including MedUSLM and the JAMA Clinical Challenge. Crucially, the patient histories in this collection average approximately 200 tokens, preserving the narrative intricacy and diagnostic ambiguity inherent in Electronic Health Records (EHRs).

\subsection{Reasoning Trace Synthesis and Filtering}
To distill expert-level cognitive capabilities, we employed a guideline-driven generation strategy utilizing GPT-5. Departing from generic generation methods, we incorporated structured medical reasoning guidelines~\cite{savage2024diagnostic} as hard constraints. Specifically, the model was instructed to synthesize reasoning traces aligned with four distinct clinical paradigms: (1) \textit{Differential Diagnosis}, (2) \textit{Intuitive Reasoning}, (3) \textit{Analytical Reasoning}, and (4) \textit{Bayesian Reasoning}. (Detailed definitions are provided in Appendix~\ref{appendix:Four distinct paradigms.}.) This ensures that the generated chains of thought reflect the diverse cognitive strategies employed by human experts, rather than formulaic textbook patterns.

To ensure data quality, we implemented a rigorous two-stage curation protocol. The first stage automatically filtered sequences based on exact-match accuracy, while the second stage involved manual verification by medical professionals to validate strict adherence to the designated reasoning paradigms. This process yielded a high-quality dataset $\mathcal{D}$ comprising $3.9$K samples, stratified into $3.3$K text-only and $0.6$K multimodal entries.

\subsection{Dataset Stratification}
Finally, we decompose the dataset $\mathcal{D}$ into four mutually exclusive subsets corresponding to distinct clinical specialties: Adult Medicine ($AM$), Child and Adolescent Health ($CAH$), Brain and Senses ($BS$), and Cancer and Surgical Interventions ($CSI$). This partition is formalized as:
\begin{equation}
    \mathcal{D} = \bigcup_{k \in \mathcal{K}} \mathcal{D}_k,
\end{equation}
where $\mathcal{K} = \{AM, CAH, BS, CSI\}$. Note that these subsets are mutually exclusive, satisfying $\mathcal{D}_i \cap \mathcal{D}_j = \varnothing$ for any $i \neq j$.

\section{MMedExpert-R1}
Current medical alignment paradigms typically rely on a two-stage framework, where a unified model undergoes Supervised Fine-Tuning (SFT) on pooled data, followed by Reinforcement Learning (RL). While superior to direct RL, which often hallucinates due to a lack of foundational knowledge, this "Single SFT+RL" approach suffers from severe parameter interference. 
Thus, we propose MMedExpert-R1, a novel decoupled multi-expert framework. Unlike prior methods that force a single parameter set to master all objectives simultaneously, we theoretically disentangle factual acquisition (via Domain-Specific Adaptation) from logical alignment (via Guideline-Based Advantages). By subsequently synthesizing these orthogonal capabilities through a Conflict-Aware Capability Integration, our approach effectively resolves the tension between broad medical knowledge and rigorous reasoning depth.




\subsection{Domain-Specific Adaptation (DSA)}
\label{sec:stage1}
To provide diverse clinical model initializations, we introduce a domain-specific adaptation.
To efficiently inject specialized knowledge without the prohibitive cost of full-parameter tuning, we employ Low-Rank Adaptation (LoRA). Let $\theta_{\text{base}}$ denote the frozen backbone parameters. For each sub-domain $k \in \{\text{AM, CAH, BS, CSI}\}$, we initialize a specific adapter $\Delta\theta_k$ and optimize it by minimizing the domain-specific SFT loss:
\begin{equation}
\Delta\theta_k^* = \operatorname*{argmin}_{\Delta\theta_k} \mathcal{L}_{\text{SFT}}(\mathcal{D}_k; \theta_{\text{base}} + \Delta\theta_k).
\end{equation}
This parallel training process yields a set of specialized parameter states, formally defined as:
\begin{equation}
\theta_k = \theta_{\text{base}} + \Delta\theta_k^*.
\end{equation}
These expert models $\theta_k$ encapsulate the distinct semantic distributions of their respective fields, providing the necessary factual basis for the subsequent integration stage.

\subsection{Guideline-Based Advantages (GBA)}
\label{sec:stage2}
Since the RL + SFT pipeline limits multi-specialty alignment due to its cold-start nature, we introduce Guideline-Based Advantage (GBA) to align optimization with the nuance of real-world clinical decision-making. Unlike standard reinforcement learning that pursues absolute rewards, GBA leverages group-relative statistics to identify and reinforce reasoning paths that best adhere to medical protocols.

For each medical query $Q$, we construct a prompt $p \in \mathcal{P}_{\text{guide}}$ containing clinical guidelines. Conditioned on $p$, we sample a group of $G$ reasoning trajectories $\{y_1, \dots, y_G\}$ from the policy $\pi_\theta$. Instead of using absolute scores, we compute the relative advantage $\mathcal{\hat{A}}_i$ for each trajectory $y_i$ by normalizing its reward against the group statistics:
\begin{equation}
\begin{split}
    \mathcal{\hat{A}}_i &= \frac{\textcolor{blue}{r(p, y_i)} - \mu_G}{\sigma_G + \epsilon}, \\
    \text{where} \quad \mu_G &= \frac{1}{G}\sum_{i=1}^G \textcolor{blue}{r(p, y_i)}.
\end{split}
\end{equation}
Here, $r(p, y_i)$ integrates diagnostic accuracy and structural constraints. Crucially, this normalization isolates the reasoning quality from the query's inherent difficulty. A positive advantage $\mathcal{\hat{A}}_i > 0$ signals superior guideline adherence relative to the group average $\mu_G$, effectively filtering out variance. Finally, we optimize the policy to maximize the likelihood of these high-advantage trajectories. The objective function uses $\mathcal{\hat{A}}_i$ to re-weight the gradient updates:
\begin{equation}
\begin{split}
\mathcal{J}_{\text{GBA}}(\theta) &= \mathbb{E}_{\substack{p \sim \mathcal{D}, \\ \{y_i\} \sim \pi_\theta}} \Biggl[ \frac{1}{G}\sum_{i=1}^G \frac{\mathcal{\hat{A}}_i}{|y_i|} \\
&\quad \cdot \sum_{t=1}^{|y_i|} \log \pi_\theta(y_{i,t} \mid p, y_{i,<t}) \Biggr].
\end{split}
\end{equation}
By explicitly penalizing deviations (where $\mathcal{\hat{A}}_i < 0$) and reinforcing valid chains, this objective shifts the policy distribution towards the guideline-compliant region.

\subsection{Conflict-Aware Capability Integration}
Since the RL with cold start limits multi-specialty alignment, we introduce Conflict-Aware Capability Integration to fuse various model experts to improve the alignment.
Specifically, we obtain two distinct categories of parameter updates: the knowledge-centric experts $\theta_{\text{AM}}$,$\theta_{\text{CAH}}$,$\theta_{\text{BS}}$,$\theta_{\text{CSI}}$ (encoding factual breadth) and the reasoning-centric model $\theta_{\text{RL}}$ (encoding logical depth). Naive averaging of these heterogeneous task vectors often precipitates catastrophic interference, where gradient drift from RL optimization overwrites the fine-grained domain features learned during SFT.
To address this, we propose a Conflict-Aware Capability Integration strategy utilizing TIES-Merging. We specifically adapt this technique to bridge the gap between distinct optimization objectives, resolving the parameter interference between factual memorization (SFT) and logical alignment (RL).

First, we isolate the capability-specific updates by transforming all fine-tuned models into a unified task vector space:
\begin{equation}
\boldsymbol{\tau}_k = \theta_k - \theta_{\text{base}},
\end{equation}
$k \in \{\text{AM, CAH, BS, CSI, RL}\}$.This separates the specific factual and reasoning injections from the base weights. Subsequently, we synthesize a robust specialist using the Trim-and-Elect mechanism:
\begin{equation}
\boldsymbol{\tau}_{\text{merged}} = \mathcal{F}_{\text{TIES}}\left( \{ \boldsymbol{\tau}_k \} ; \lambda \right).
\end{equation}
The operator $\mathcal{F}_{\text{TIES}}$ filters stochastic noise by retaining only the top-$k\%$ magnitude parameters and enforcing sign consistency. This ensures that only consensual features—validated across multiple domains or reinforced by reasoning alignment—are preserved. The final dual-stream agent is reconstructed via $\theta_{\text{merged}} = \theta_{\text{base}} + \eta \cdot \boldsymbol{\tau}_{\text{merged}}$, effectively harmonizing granular domain knowledge with rigorous clinical logic.

\begin{table*}[t]
\centering
\caption{Performance comparison of MMedExpert-R1 with other MedVLMs on medical multimodal benchmarks, where bold indicates the best. Note that OMVQA indicates OmniMedVQA. MedVLM-R1 and Med-R1 is trained on part of the OmniMedVQA test set, making its results on OmniMedVQA meaningless.}
\label{tab:model_comparison}
\resizebox{\textwidth}{!}{
\setlength{\tabcolsep}{2.5pt}
\renewcommand{\arraystretch}{1.2}
\begin{tabular}{@{}llcccccccccc@{}}
\toprule
\multirow{2}{*}{\textbf{Base}} & \multirow{2}{*}{\textbf{Size}} & \multicolumn{6}{c}{\textbf{MedXpert-MM}} & \multirow{2}{*}{\textbf{PMC-VQA}} & \multirow{2}{*}{\textbf{OMVQA}} &\multirow{2}{*}{\textbf{GMAI-MM}} \\
\cmidrule(lr){3-8}
 & & \textbf{Total} & \textbf{Rea.} & \textbf{Under.} & \textbf{Treatment} & \textbf{Basic Science} & \textbf{Diagnosis} &  & \\
\midrule
\rowcolor{gray!5} MedVLM-R1 & 2B & 19.95 & 19.84 & 20.21 & 20.98 &17.28 & 20.35 & 47.83 & - &41.18 \\
\rowcolor{gray!5} Med-R1 & 2B & 21.15 & 20.88 & 21.84 & 24.33 &17.28 & 21.10 & 45.48 & - &40.79 \\
\rowcolor{gray!5} MedVLThinker & 3B & 21.90 & 20.80 & \textbf{24.72} & 23.66 
&\textbf{20.11} & 21.76 & 50.95 & 64.17 &38.81 \\
\rowcolor{green!10} MMedExpert-R1 &2B &\textbf{23.35} &\textbf{23.02} 	&24.18 	&\textbf{28.12} 	&19.54	&\textbf{22.68} 		&\textbf{53.27} &\textbf{64.64} & \textbf{43.67}
 \\ \hline
\rowcolor{gray!5} LLaVA-Med & 7B & 19.65 & 19.43 & 20.21 & 22.32 &19.54 & 18.68 & 30.07 & 43.68 &31.23 \\
\rowcolor{gray!5} HuatuoGPT-V & 7B & 22.65 & 21.43 & 25.81 & 21.65 &20.11 & 23.76 & 53.40 & 75.46 &51.27\\
\rowcolor{gray!5} BiMediX2 & 7B & 22.15 & 21.23 & 24.54 & 22.54 &21.24 & 22.26 & 42.79 & 62.42 &34.59 \\
\rowcolor{gray!5} Lingshu & 7B & 26.00 & 25.44 & 27.43 & 28.12 &22.09  & 26.35 & 54.68 & 82.85  &51.21\\ 
\rowcolor{green!10} MMedExpert-R1 &7B  &\textbf{27.50} 	&\textbf{26.55} 	&\textbf{29.96} 	&\textbf{29.91} 	&\textbf{22.66} 	&\textbf{28.02} 		&\textbf{56.78} & \textbf{83.03}&\textbf{52.10} \\
\bottomrule
\end{tabular}}
\end{table*}

\section{Experiments}
\subsection{Experiment Setting}

To comprehensively evaluate the performance of MMedExpert-R1 across a broad range of clinical reasoning tasks, we benchmark it against a diverse set of state-of-the-art medical vision-language models, including MedVLM-R1~\cite{pan2025medvlm}, Med-R1~\cite{lai2025med}, MedVLThinker~\cite{huang2025medvlthinker}, MedGemma~\cite{sellergren2025medgemma}, LLaVA-Med~\cite{li2023llava}, HuatuoGPT-V~\cite{chen2024huatuogpt}, BiMediX2~\cite{mullappilly2024bimedix2}, and Lingshu~\cite{xu2025lingshu}. To ensure fair and consistent comparisons, all models are assessed within a standardized environment using our proposed MedEvalKit. This framework unifies experimental settings, prompts, and evaluation metrics across all candidate models.

For benchmark datasets, we utilize the test sets of \textbf{PMC-VQA (v2)}~\cite{zhang2023pmc}, \textbf{OmniMedVQA}~\cite{hu2024omnimedvqa}, \textbf{GMAI-MMBench (val)}~\cite{ye2024gmai}, and \textbf{MedXpertQA-MM}~\cite{zuo2025medxpertqa}. Collectively, these benchmarks cover a comprehensive spectrum of medical imaging modalities, spanning X-ray, CT, MRI, PET, ultrasound, microscopy, pathology, OCT, dermoscopy, gastrointestinal (GI) examinations, endoscopy, and fundus imaging, as well as medical charts, tables, and figures. This extensive coverage ensures a rigorous and holistic assessment of multimodal medical reasoning capabilities.
Implementation Details are shown in the Appendix~\ref{appendix:Implementation Details}.

\begin{table}[t]
\centering
\caption{Results on the validation set of GMAI-MMBench for clinical VQA tasks. CR denotes Cell Recognition, SAR denotes Surgeon Action Recognition, OR-T denotes Organ Recognition - Thorax, OR-H denotes Organ Recognition - Head and Neck, OR-A denotes Organ Recognition - Abdomen, NT denotes Nervous Tissue, MR denotes Microorganism Recognition. } 
\label{tab:gmaimmbench_optimized}
\footnotesize 
\renewcommand{\arraystretch}{1.3} 
\setlength{\tabcolsep}{2pt} 

\scalebox{0.9}{ 
\begin{tabular}{@{}lccccccc@{}}
\toprule
Base & CR & SAR & OR-T &OR-H & OR-A &NT &MR  \\
\midrule
\rowcolor{gray!5} MedVLM-R1 &41.73 &26.08 &42.35 &41.29 &36.32 &57.50 &36.29 \\
\rowcolor{gray!5} Med-R1 &42.60 &27.82 &40.00 &40.00 &35.91 &65.00 &37.03 \\
\rowcolor{gray!5} MedVLThinker &40.86 &27.82 &38.23 &43.87 &\textbf{43.67} &52.50 &30.37 \\
\rowcolor{green!10} MMedExpert-R1 &\textbf{42.60} &\textbf{28.69} &\textbf{42.94} &\textbf{50.32} &34.28 &\textbf{72.50} &\textbf{39.25} \\\hline
\rowcolor{gray!5} MedGemma &31.30 &24.34 &51.17 &57.41 &42.44 &55.00 &35.55 \\ 
\rowcolor{gray!5} LLaVA-Med &24.34 &13.04 &24.11 &20.64 &31.85 &47.50 &34.81  \\
\rowcolor{gray!5} HuatuoGPT-V &38.26 &22.60 &56.47 &60.00 &49.79 &72.50 &51.11 \\
\rowcolor{gray!5} BiMediX2 &27.49 &24.34 &33.52 &37.41 &28.57 &35.00 &35.55  \\
\rowcolor{gray!5} Lingshu &39.13 &27.82 &51.17 &65.16 &50.10 &77.50 &57.03 \\
\rowcolor{green!10} MMedExpert-R1 &\textbf{40.00} &\textbf{31.30} &\textbf{60.58} &\textbf{65.16} &\textbf{54.28} &\textbf{80.00} &\textbf{57.77}  \\
\bottomrule
\end{tabular}}
\end{table}




\subsection{Experimental Results}

We evaluate MMedExpert-R1 against MedVLMs across varying parameter scales (2B and 7B).

\subsubsection{Main Results and Scaling Analysis} 

As presented in Table~\ref{tab:model_comparison}, MMedExpert-R1 demonstrates superior overall performance compared to state-of-the-art MedVLMs across both the 2B and 7B parameter scales.
In the lightweight group ($<4$B parameters), MMedExpert-R1 (2B) establishes a distinct advantage across multiple benchmarks. It achieves a Total score of $23.35$ on MedXpert-MM, surpassing the 3B-scale MedVLThinker ($21.90$). This superiority extends to external benchmarks, where our 2B model records $53.27$ on PMC-VQA and $43.67$ on GMAI-MM, significantly outperforming MedVLM-R1 ($47.83$ and $41.18$).
Upon scaling to 7B parameters, MMedExpert-R1 achieves comprehensive state-of-the-art results. It secures a leading score of $27.50$ on MedXpert-MM, while also dominating public benchmarks with scores of $56.78$ on PMC-VQA, $83.03$ on OMVQA, and $52.10$ on GMAI-MM, consistently exceeding strong competitors like Lingshu ($54.68$, $82.85$, and $51.21$) and HuatuoGPT-V.

\subsubsection{Fine-grained Analysis of Clinical Capabilities}
To further investigate the source of these performance gains, we analyze the model's proficiency in specific clinical tasks using metrics from both MedXpert-MM (Table~\ref{tab:model_comparison}) and the GMAI-MM validation set (Table~\ref{tab:gmaimmbench_optimized}). 

Results from MedXpert-MM highlight MMedExpert-R1's exceptional capability in high-stakes clinical decision-making. Specifically, our 2B model achieves a "Treatment" score of $28.12$, dominating the 2B/3B tier, while the 7B model leads the "Diagnosis" task with a score of $28.02$. 
This clinical expertise is further corroborated by the granular breakdown in Table 2. MMedExpert-R1 (7B) demonstrates robust capabilities across diverse visual recognition tasks, achieving the highest scores in Cell Recognition (CR) with $40.00$ and Microorganism Recognition (MR) with $57.77$. Furthermore, it excels in anatomical localization, leading in Organ Recognition - Head and Neck (OR-H) ($65.16$) and Organ Recognition - Abdomen (OR-A) ($54.28$). Even the 2B variant displays competitive performance, particularly in Nervous Tissue (NT) recognition ($72.50$) and Surgeon Action Recognition (SAR) ($28.69$). These results indicate that our expert-guided alignment strategy effectively enhances the model's ability to handle complex, domain-specific visual challenges ranging from microscopic identification to surgical context understanding. (More results are shown in the Appendix~\ref{appendix: gmai results} .)


\begin{figure*}[t]
  \includegraphics[width=\textwidth]{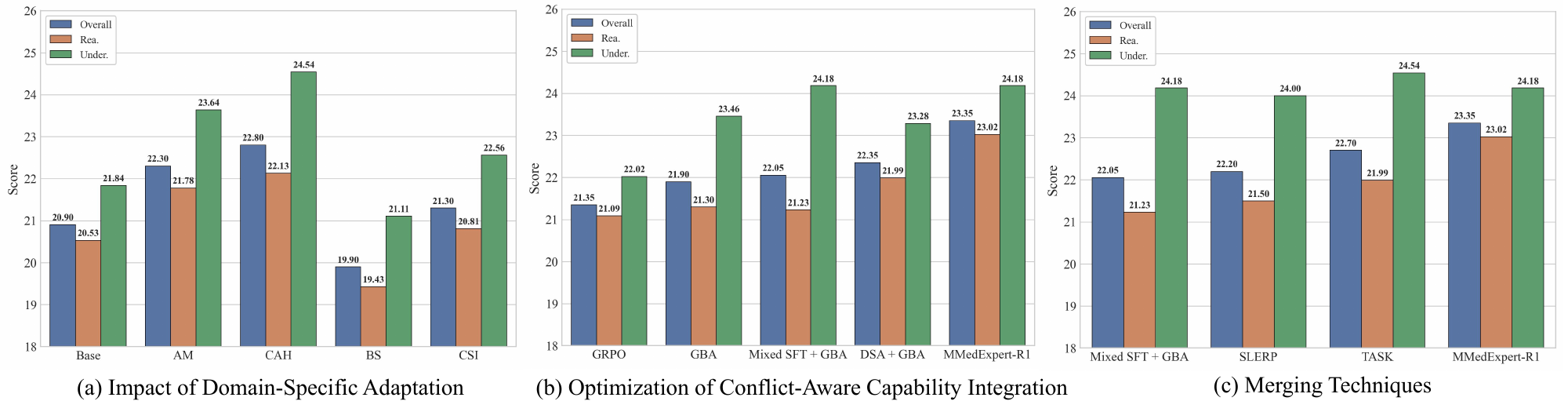}
  \caption{ Ablation studies on MMedExpert-R1. (a) illustrates the impact of Domain-Specific Adaptation. (b) shows optimization of Conflict-Aware Capability Integration. (c) demonstrates the superiority of our merging technique.
}
  \label{fig:ablation_main}
\end{figure*}

\subsubsection{Ablation Study}

\begin{table}[t]
\centering
\caption{Ablation study of different training strategies and components in MMedExpert-R1. \textbf{Unified}: Training on pooled data. \textbf{Disentangled}: Training specialized experts. \textbf{TIES}: Conflict-aware merging.}
\label{tab:ablation_study_setting}
\footnotesize
\renewcommand{\arraystretch}{1.2}
\setlength{\tabcolsep}{3pt}
\scalebox{0.7}{
\begin{tabular}{@{}l|cc|cc|cc|c@{}} 
\toprule
\multirow{2}{*}{\textbf{Method}} & \multicolumn{2}{c|}{\textbf{DSA}} & \multicolumn{2}{c|}{\textbf{Advantage}} & \multicolumn{2}{c|}{\textbf{CACI}} & \multirow{2}{*}{\textbf{Avg. Score}} \\ \cmidrule(lr){2-3} \cmidrule(lr){4-5} \cmidrule(lr){6-7}
 & Unified & Disentangled & GRPO & GBA & Naive & TIES & \\ \midrule
Base & - & - & - & - & - & - & 20.90 \\ \midrule
AM & - & \checkmark (AM) & - & - & - & - & 22.30 \\
CAH & - & \checkmark (CAH) & - & - & - & - & 22.80 \\ 
BS & - & \checkmark (BS) & - & - & - & - & 19.90 \\ 
CSI & - & \checkmark (CSI) & - & - & - & - & 21.30 \\ 
Mixed SFT & \checkmark & - & - & - & - & - & 22.05 \\ \midrule
GRPO & - & - & \checkmark & - & - & - & 21.35 \\
GBA & - & - & - & \checkmark & - & - & 21.90 \\ \midrule
Mixed SFT + GBA & \checkmark & - & - & \checkmark & \checkmark & - & 22.05 \\
DSA + GBA & - & \checkmark & - & \checkmark & \checkmark & - &22.35 \\
\textbf{Ours} & - & \checkmark & - & \checkmark & - & \checkmark & \textbf{23.35} \\ \bottomrule
\end{tabular}}
\end{table}

We conduct comprehensive ablation studies to investigate the effectiveness of each component in MMedExpert-R1. The experiments are categorized into three aspects: the impact of Domain-Specific Adaptation, the effectiveness of Reasoning-Specific Adaptation, and the optimization of Conflict-Aware Capability Integration and merging methods. Detailed categories are shown in Table~\ref{tab:ablation_study_setting}.

\noindent\textbf{Impact of Domain-Specific Adaptation.}
We first evaluate the contribution of the Domain-Specific Adaptation (DSA) stage by training individual experts on different data categories (denoted as AM, CAH, BS, and CSI). 
As shown in Fig~\ref{fig:ablation_main} (a), models trained on specific high-quality subsets generally outperform the Base model. Notably, CAH achieves a significant boost, reaching an overall score of $22.80$ compared to the Base score of $20.90$. It is observed that while some domains like BS may not immediately yield general improvements due to domain shifts, the strong performance of CAH and AM validates our hypothesis: even with a limited amount of data, high-quality, domain-specific instruction tuning can yield substantial performance gains. This justifies our strategy of training distinct experts before aggregation.

\noindent\textbf{Effectiveness of Guideline-Based Advantages.}
In the second block of Fig~\ref{fig:ablation_main} (b), we compare the efficacy of standard Group Relative Policy Optimization (GRPO) against our proposed Guideline-Based Advantage (GBA). The results demonstrate that GBA ($21.90$) consistently outperforms the primitive GRPO ($21.35$) across all metrics. This improvement suggests that incorporating multi-view perspectives during the alignment phase enables the model to capture more robust feedback gradients, thereby enhancing its reasoning and understanding capabilities more effectively than standard policy optimization methods.




\noindent\textbf{Optimization of Conflict-Aware Capability Integration.}
We further analyze different strategies for integrating the DSA and GBA. We compare three settings: Mixed SFT + GBA: Mixing all expert data during the DSA stage, followed by GBA training. DSA + GBA: Merging expert models after the DSA stage, followed by GBA training. MMedExpert-R1 (Ours): A multi-stage approach where we perform model merging after both the DSA and GBA stages. As shown in the bottom block of Table 3, the mixed data training (Mixed SFT + GBA) yields limited improvement ($22.05$). While merging after DSA (DSA + GBA) improves the score to $22.35$, our full multi-stage merging strategy (MMedExpert-R1) achieves the highest performance of $23.35$. This indicates that preserving expert specialization through separate training and merging them at multiple stages is superior to simply mixing data or merging only once.

\noindent\textbf{Analysis of Merging Techniques.}
Finally, we investigate the impact of different model merging algorithms on the final performance. Based on the optimized pipeline, we compare our method against standard merging techniques, including Spherical Linear Interpolation (SLERP) and Task Arithmetic (TASK). As presented in Fig~\ref{fig:ablation_main} (c), while Task Arithmetic ($22.70$) outperforms SLERP ($22.20$) and the mixed baseline, MMedExpert-R1 achieves the best results ($23.35$). This confirms that our merging strategy effectively retains the distinct capabilities of each expert model without suffering from parameter interference, leading to the most robust aggregated model.

\section{Conclusion}

In this paper, we presented MMedExpert-R1, a novel MedVLM framework that effectively bridges the gap between limited high-quality data and robust clinical reasoning capabilities. By introducing a pipeline that integrates Domain-Specific Adaptation for expert specialization and Guideline-Based Advantages for precise preference optimization via advantages based on different clinical guidelines, we successfully synthesized a model that excels in complex medical scenarios. Our comprehensive evaluation across four major benchmarks demonstrates that MMedExpert-R1 achieves state-of-the-art performance at both the 2B and 7B parameter scales.

\section*{Limitations}
Although our framework demonstrates strong performance across standard multimodal medical benchmarks, there are limitations that are practical to address in near-term follow-up work. 
First, our expert specialization uses four broad LoRA domains (Adult Medicine, Child and Adolescent Health, Brain and Senses, Cancer and Surgical Interventions); introducing finer-grained subdomains (e.g., cardiology, pulmonology, neurology subspecialties) is a straightforward extension that may further improve specialization without architectural changes. Second, although we evaluate on widely used multimodal medical benchmarks, broadening the evaluation suite to include additional datasets and task variants (e.g., more ophthalmology/dermatology sets, longitudinal comparisons, report summarization) using our MedEvalKit would provide more comprehensive validation.



\bibliography{custom}

\appendix
\clearpage
\textbf{\large Appendix for MMedExpert-R1}

\vspace{10pt}
\noindent\textbf{Abstract.}
\label{sec:appendix}

Appendix~\ref{appendix:Four distinct paradigms.} shows the explanation for "four distinct paradigms: (1) differential diagnosis, (2) intuitive reasoning, (3) analytical reasoning, and (4) Bayesian reasoning".

Appendix~\ref{appendix:Implementation Details} details the training hyperparameters.

Appendix~\ref{appendix: gmai results} shows the complete results on GMAI-MMBench and OmniMedVQA.

Appendix~\ref{appendix:Visualization results} shows the visualization results on the MedXpert-MM benchmark.

Appendix~\ref{appendix: Computational Efficiency Analysis} details the cost of computation.

\section{Four Distinct Paradigms}
\label{appendix:Four distinct paradigms.}

To clarify how the four paradigms guide distinct reasoning trajectories for the same case, we provide the following definitions and structured examples:

\vspace{0.5em} 

\paragraph{Differential Diagnosis Reasoning.} 
Construct a differential diagnosis list, compare competing etiologies step by step, and eliminate them using imaging and clinical cues. 

\noindent\textit{Example:} (1) The lesion is round and circumscribed, favoring benignity. (2) Phyllodes tumor: typically larger, lobulated. (3) Invasive ductal carcinoma with DCIS: irregular, spiculated margins---absent here. (4) Fibroadenoma: smooth, encapsulated mass---matches best.

\paragraph{Intuitive Reasoning.} 
Apply pattern recognition and association of visual features with typical diseases. 

\noindent\textit{Example:} (1) Smooth, oval mass with clear borders is intuitively recognized as benign. (2) Absence of calcifications or distortion supports nonmalignant nature. (3) Common benign lesion in this pattern is fibroadenoma.

\paragraph{Analytical Reasoning.} 
Use mechanistic and pathophysiologic deduction based on structural and cellular processes. 

\noindent\textit{Example:} (1) Malignancies induce stromal desmoplasia $\rightarrow$ spiculation (absent). (2) Intraductal lesions form cystic or complex masses (not observed). (3) Metastatic nodules are usually multiple. (4) Benign stromal proliferation forming a smooth, solid encapsulated nodule fits fibroadenoma.

\paragraph{Bayesian Reasoning.} 
Start from prior probabilities, update with new imaging evidence, and infer posterior likelihoods. 

\noindent\textit{Example:} (1) Prior: Fibroadenoma most common benign mass in screening-age group. (2) Evidence: Smooth border increases $P(\text{benign})$; absence of distortion decreases $P(\text{malignant})$. (3) Posterior: Highest posterior probability assigned to fibroadenoma after evidence update.

\section{Implementation Details}
\label{appendix:Implementation Details}
Our experiments are conducted using Qwen2-VL-2B-Instruct and Lingshu, the multimodal Large Vision-Language Model (LVLM) equipped with a pretrained visual encoder capable of handling image resolutions up to 401,408 pixels. 

For the DSA stage, each domain expert is optimized using Low-Rank Adaptation (LoRA). We apply LoRA to all linear layers with a rank $r=16$ and scaling factor $\alpha=32$, utilizing bfloat16 precision. The training process spans 3 epochs with a learning rate of $1 \times 10^{-6}$, a batch size of 1, and 8 gradient accumulation steps.

For the Reinforcement Learning (RL) stage, we set group size $G=8$, clipping parameter $\epsilon=0.2$, and KL penalty coefficient $\beta=0.001$. The maximum rollout length is set to 1024. To ensure computational efficiency during distributed training, we utilize DeepSpeed ZeRO-2 optimization.


\section{Complete Results on GMAI-MMBench}
\label{appendix: gmai results}

Table~\ref{tab:gmaimmbench_appendix} shows the complete results on GMAI-MMBench.  Meanwhile, we trained MMedExpert-R1 using a subset of 1000 OmniMedVQA dataset, and the results showed a significant improvement, even surpassing the performance of MedVLM-R1 trained with 80\% of the data, as shown in Table~\ref{tab:omnimedvqa_results}.

\begin{table}[htbp]
  \centering
  \caption{Comparison of different MLLMs on OmniMedVQA benchmark.}
  \label{tab:omnimedvqa_results}
  \begin{tabular}{lc}
    \toprule
    \textbf{Model} & \textbf{OmniMedVQA} \\
    \midrule
    MedVLM-R1      & 77.38 \\
    Med-R1         & 76.08 \\
    MMedExpert-R1  & \textbf{77.93} \\
    \bottomrule
  \end{tabular}
\end{table}

\begin{table*}[t]
\centering
\caption{Results on the validation set of GMAI-MMBench for clinical VQA tasks.} 
\label{tab:gmaimmbench_appendix}
\footnotesize 
\renewcommand{\arraystretch}{1.3} 
\setlength{\tabcolsep}{2pt} 

\scalebox{0.85}{ 
\begin{tabular}{@{}llcccccccccccccccccc@{}}
\toprule
\textbf{Base} & \textbf{Total} & \textbf{DD} & \textbf{SG} & \textbf{CR} & \textbf{IQG} & \textbf{SAR} & \textbf{OR-T} & \textbf{BVR} & \textbf{B} & \textbf{OR-H} & \textbf{OR-A} & \textbf{NT} & \textbf{MR} & \textbf{SR} & \textbf{AR} & \textbf{OR-P} & \textbf{M} & \textbf{SWR} & \textbf{C} \\
\midrule
\rowcolor{gray!5} MedVLM-R1 & 41.18 & 48.16 & 23.01 & 41.73 & 32.00 & 26.08 & 42.35 & \textbf{44.44} & \textbf{40.00} & 41.29 & 36.32 & 57.50 & 36.29 & 30.95 & 42.96 & \textbf{42.66} & \textbf{32.00} & 31.42 & 23.40 \\
\rowcolor{gray!5} Med-R1 & 40.79 & 47.45 & 25.52 & \textbf{42.60} & \textbf{34.00} & 27.82 & 40.00 & 40.74 & 38.85 & 40.00 & 35.91 & 65.00 & 37.03 & \textbf{31.84} & \textbf{45.92} & 30.66 & 24.00 & 34.28 & 23.93 \\
\rowcolor{gray!5} MedVLThinker & 38.81 & 43.16 & \textbf{26.35} & 40.86 & 26.00 & 27.82 & 38.23 & 42.96 & 37.14 & 43.87 & \textbf{43.67} & 52.50 & 30.37 & 30.95 & 36.29 & 20.00 & 30.00 & \textbf{37.14} & \textbf{32.44} \\
\rowcolor{green!10} MMedExpert-R1 &\textbf{43.67} &\textbf{52.02} &25.94 &\textbf{42.60} &\textbf{34.00} &\textbf{28.69} &\textbf{42.94}&42.96 & 38.28 &\textbf{50.32} &34.28 &\textbf{72.50} &\textbf{39.25}&31.54 &42.96 &40.00 &26.00 &31.42 &27.12\\\hline
\rowcolor{gray!5} MedGemma & 45.25 & 52.73 & 25.10 & 31.30 & 30.00 & 24.34 & 51.17 & 55.55 & 49.71 & 57.41 & 42.44 & 55.00 & 35.55 & 33.63 & 46.66 & 41.33 & 30.00 & 34.28 & 22.87 \\ 
\rowcolor{gray!5} LLaVA-Med & 31.23 & 36.05 & 23.01 & 24.34 & 18.00 & 13.04 & 24.11 & 25.18 & 22.85 & 20.64 & 31.85 & 47.50 & 34.81 & 37.50 & 37.03 & 17.33 & 20.00 & 24.28 & 22.34 \\
\rowcolor{gray!5} HuatuoGPT-V & 51.27 & \textbf{59.70} & 29.70 & 38.26 & \textbf{44.00 }& 22.60 & 56.47 & \textbf{60.74} & 49.71 & 60.00 & 49.79 & 72.50 & 51.11 & 32.14 & \textbf{49.62} & 49.33 & 30.00 & \textbf{38.57} & \textbf{37.76} \\
\rowcolor{gray!5} BiMediX2 & 34.59 & 39.34 & 24.26 & 29.47 & 28.00 & 24.34 & 33.52 & 36.29 & 26.85 & 37.41 & 28.57 & 35.00 & 35.55 & \textbf{33.63} & 36.29 & 37.33 & 34.00 & 35.71 & 19.68 \\
\rowcolor{gray!5} Lingshu & 51.21 & \textbf{59.70} & 26.77 & 39.13 & 40.00 & 27.82 & 51.17 & 54.07 & 52.71 & \textbf{65.16} & 50.10 & 77.50 & 57.03 & 31.84 & 45.18 & 60.66 & 40.00 & 28.57 & 32.44 \\
\rowcolor{green!10} MMedExpert-R1  &\textbf{52.10} &58.71 &\textbf{30.96}&\textbf{40.00} &40.00 &\textbf{31.30} &\textbf{60.58} &52.59 &\textbf{60.00} &\textbf{65.16} &\textbf{54.28} &\textbf{80.00} &\textbf{57.77} &32.44 &47.40 &\textbf{70.66} &\textbf{42.00} &31.42 &30.31 \\
\bottomrule
\end{tabular}}
\end{table*}

\section{Visualization Results}
\label{appendix:Visualization results}

\begin{figure*}[t]
  \includegraphics[width=\textwidth]{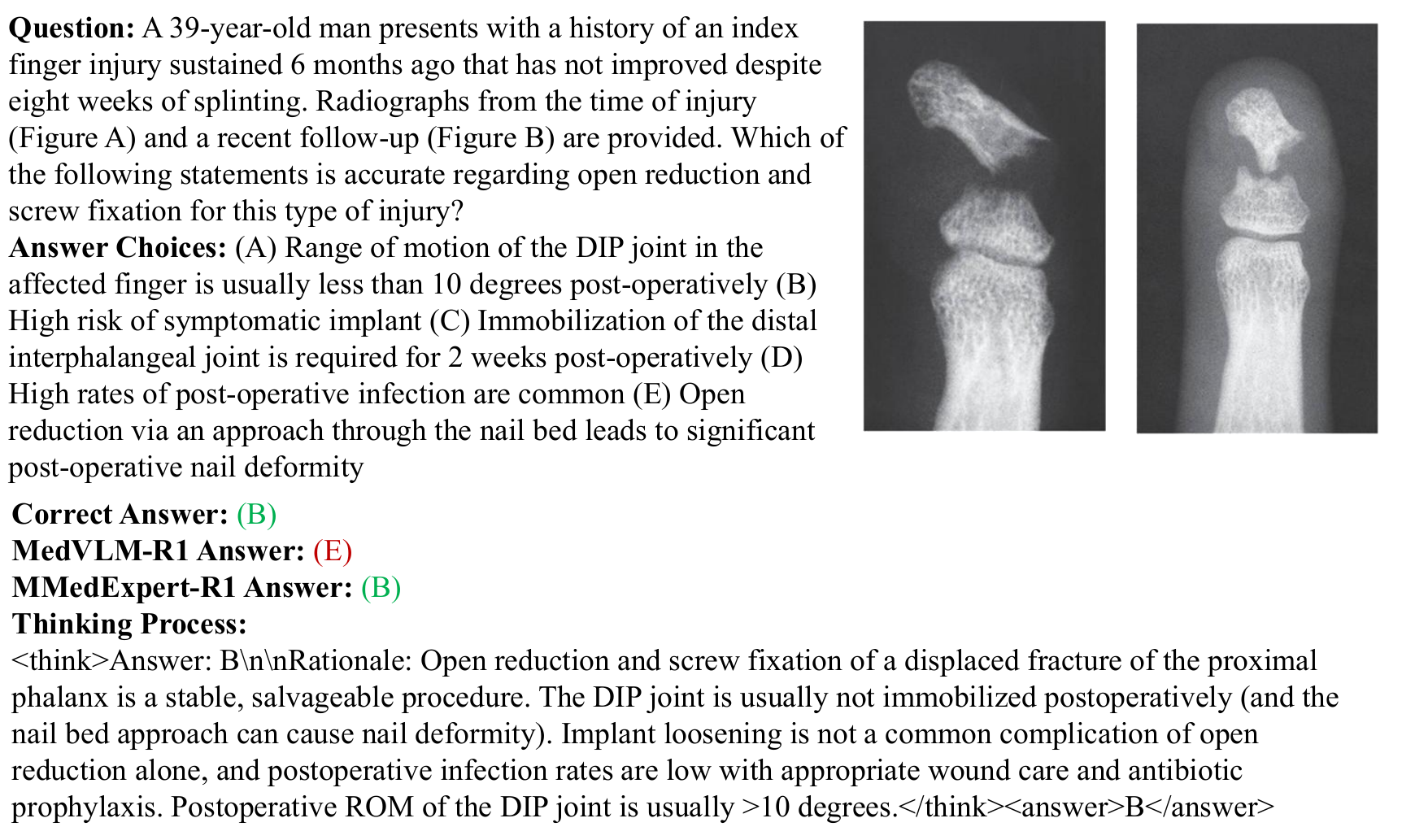}
  \caption{Visualization on MedXpert-MM.
}
  \label{fig:example1}
\end{figure*}

\begin{figure*}[t]
  \includegraphics[width=\textwidth]{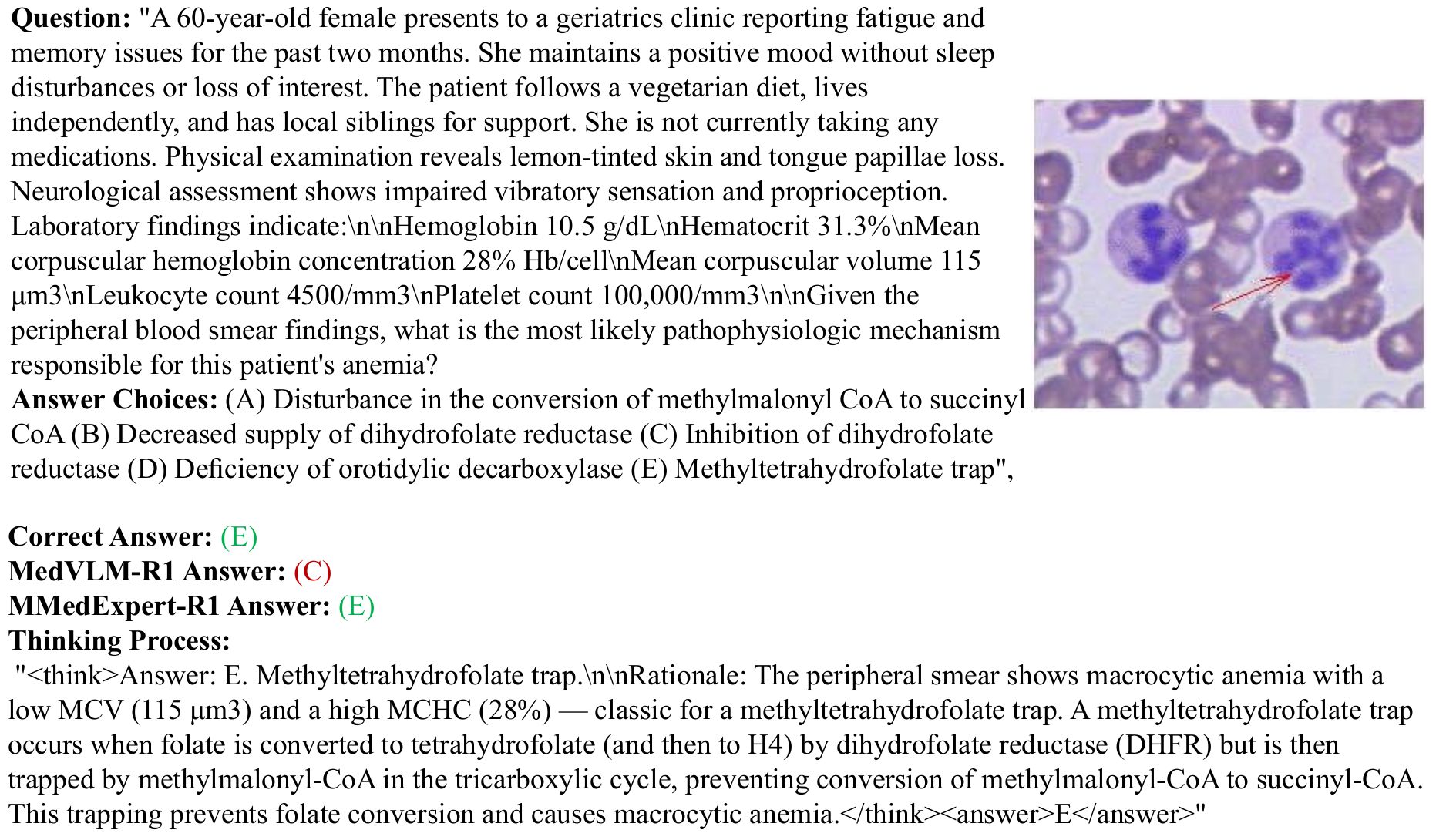}
  \caption{Visualization on MedXpert-MM.
}
  \label{fig:example2}
\end{figure*}

Fig~\ref{fig:example1} and Fig~\ref{fig:example2} show the visualization results on the MedXpert-MM benchmark.

\section{Computational Efficiency Analysis}
\label{appendix: Computational Efficiency Analysis}

We provide detailed information on computational cost and latency across all major pipeline components to demonstrate deployment feasibility. All estimates are based on experiments conducted on a single NVIDIA A100 GPU.
\paragraph{Domain-Specific Adaptation (DSA).} Leveraging the lightweight nature of LoRA-based fine-tuning, training on approximately 1,000 samples requires only $\approx$0.4 hours. This low overhead facilitates rapid adaptation to new domains.
\paragraph{Guideline-Based Advantages (GBA).} RL fine-tuning is the most compute-intensive stage, requiring $\approx$13 hours for 1,000 samples. The increased computational cost is primarily driven by the need for multi-sample generation per iteration and the calculation of KL-divergence for policy regularization.
\paragraph{Conflict-Aware Capability Integration.} The merging of domain experts is highly efficient, taking $\approx$5 minutes per merge operation. Crucially, this process results in a single consolidated model, incurring \textbf{negligible additional latency} during inference compared to a standard base model.




\end{document}